%% file: aaai25.tex
\documentclass[letterpaper]{article} 
\usepackage{aaai25}  
\usepackage{times}  
\usepackage{helvet}  
\usepackage{courier}  
\usepackage[hyphens]{url}  
\usepackage{graphicx} 
\usepackage{natbib}  
\usepackage{caption} 
\usepackage{algorithm}
\usepackage{algorithmic}

\usepackage{booktabs}
\usepackage{appendix}
\usepackage{subfigure}
\usepackage{listings}
\usepackage{xcolor}
\usepackage{graphicx}
\usepackage{makecell}
\usepackage{subfig}
\usepackage{multirow}
\usepackage{enumitem}
\usepackage{amsmath, amssymb, amsfonts}
\usepackage{utfsym}
\usepackage{threeparttable}
\usepackage{bm}

\newcommand{\repo}{\url{https://github.com/kg-cc/NoiseHGNN}}

\usepackage{newfloat}
\usepackage{listings}

\DeclareCaptionStyle{ruled}{labelfont=normalfont,labelsep=colon,strut=off} 
\floatstyle{ruled}
\newfloat{listing}{tb}{lst}{}
\floatname{listing}{Listing}
\title{NoiseHGNN: Synthesized Similarity Graph-Based Neural Network For Noised Heterogeneous Graph Representation Learning}
\author{
    Xiong Zhang\textsuperscript{\rm 1}, Cheng Xie\textsuperscript{\rm 1}\thanks{Corresponding author.}
    Haoran Duan\textsuperscript{\rm 2} , Beibei Yu\textsuperscript{\rm 3}  }
\affiliations{

    \textsuperscript{\rm 1} School of Software, Yunnan University,  Kunming,  China\\
	\textsuperscript{\rm 2} School of Cyber Science and Engineering, Wuhan University, Wuhan, China \\
	\textsuperscript{\rm 3} Australian AI Institute, University of Technology Sydne, Sydney, Australia \\
        
        \{zhangxiong@stu, xiecheng\}@ynu.edu.cn, hrduan07@gmail.com, Beibei.Yu@student.uts.edu.au \\

}

\usepackage{bibentry}

\begin{document}

\maketitle

\begin{abstract}
Real-world graph data environments intrinsically exist noise (e.g., link and structure errors) that inevitably disturb the effectiveness of graph representation and downstream learning tasks.
For homogeneous graphs, the latest works use original node features to synthesize a similarity graph that can correct the structure of the noised graph.
This idea is based on the homogeneity assumption, which states that similar nodes in the homogeneous graph tend to have direct links in the original graph.
However, similar nodes in heterogeneous graphs usually do not have direct links, which can not be used to correct the original noise graph. 
This causes a significant challenge in noised heterogeneous graph learning.
To this end, this paper proposes a novel synthesized similarity-based graph neural network compatible with noised heterogeneous graph learning.
First, we calculate the original feature similarities of all nodes to synthesize a similarity-based high-order graph.
Second, we propose a similarity-aware encoder to embed original and synthesized graphs with shared parameters.
Then, instead of graph-to-graph supervising, we synchronously supervise the original and synthesized graph embeddings to predict the same labels.
Meanwhile, a target-based graph extracted from the synthesized graph contrasts the structure of the metapath-based graph extracted from the original graph to learn the mutual information.
Extensive experiments in numerous real-world datasets show the proposed method achieves state-of-the-art records in the noised heterogeneous graph learning tasks.
In highlights, +5$\sim$6\% improvements are observed in several noised datasets compared with previous SOTA methods. 
The code and datasets are available at \repo.
\end{abstract}

\section{Introduction}
\label{sec:introduction}

\input{section/01_Introdiction}

\label{sec:Related_Work}
\input{section/03_Related_Work}

\label{sec:Preliminaries}
\input{section/02_Preliminaries}

\label{sec:Proposed_Method}
\input{section/04_Proposed_Method}

\label{sec:Experiments}
\input{section/05_Experiments}

\label{sec:conclusion}
\input{section/06_Conclusion}

\section{Acknowledgments}
This paper is the result of the research project funded by the National Natural Foundation of China (Grant No. 62106216 and 62162064) and the Open Foundation of Yunnan Key Laboratory of Software Engineering under Grant No.2023SE104.

\bibliography{aaai25}

\clearpage
\input{section/07_appendix}

\end{document}

%% file: section/01_Introdiction.tex
Graph representation learning is one of the most significant research fields in artificial intelligence, since most intelligent applications are based on graph representations such as recommendation systems \cite{2_lv2021we,4_fan2019graph,5_zhao2017meta}, social networks \cite{11_qiu2018deepinf,12_li2019encoding,14_wang2019online}, biomedicine \cite{6_gaudelet2021utilizing,13_fout2017protein,17_davis2019comparative}, fraud detection \cite{7_shchur2018pitfalls,9_dou2020enhancing}, e-commerce \cite{15_ji2021large,16_zhao2019intentgc}, etc.
However, real-world graph data environments inherently contain noise data (e.g., structure errors, etc.) that challenge the representation models. Specifically, representation models are required to maintain their effectiveness in noised graph data environments.

\begin{figure}[!t]
    \centering
    \includegraphics[width=1\linewidth]{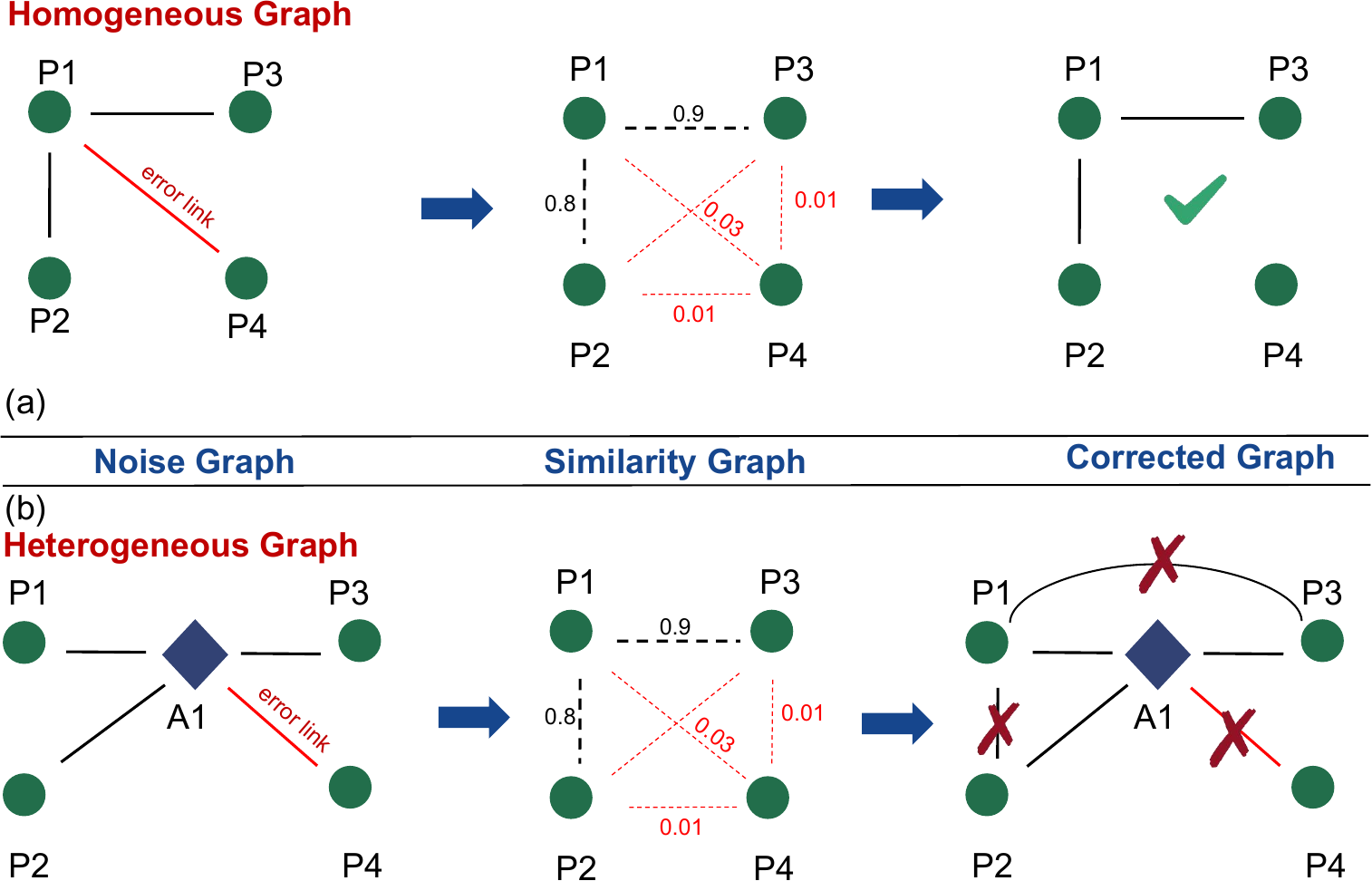}
    \caption{Homogeneity assumption works for noised homogeneous graph but failed in noised heterogeneous graph.}
    \label{fig: motivation}
\end{figure}



The latest works attempting to solve this problem by optimizing graph structure  \cite{30_wang2021graph, wei2022contrastive} or synthesizing graph structure \cite{39_fatemi2021slaps,40_franceschi2019learning,41_jin2020graph,42_liu2022towards} to alleviate the noise.
Nowadays, graph synthesizing-based methods achieve impressive performance in noised homogeneous graph representation learning, based on the homogeneity assumption to create a synthesized graph to correct the noise homogeneous graph.
As an example shown in Fig.\ref{fig: motivation} (a), a paper citation graph has four paper nodes: P1, P2, P3, and P4.  
The links between nodes represent the references among the papers.
The red link means P1 and P4 are incorrectly realized to have a reference relationship.
To fix these error links, a similarity graph can be constructed in advance by calculating the node-to-node feature similarities without using links.
It can be observed that P1 and P4 have minimal similarity values (0.03), which means P1 is unlikely to cite P4 in the original graph.
Thus, the original noised graph will likely be corrected by calculating the loss between it and the similarity graph.
This idea is based on the homogeneity assumption, which states that similar nodes tend to have direct links in a homogeneous graph.

However, the above solution can not be simply applied to the heterogeneous graphs.
As an example shown in Fig.\ref{fig: motivation} (b), a heterogeneous graph represents the relationships between papers and authors. 
A similarity graph shows that P1, P2, and P3 are similar enough to have a link.
We will obtain a wrong correction result if we directly calculate the loss between the original and similarity graphs.
This is because links in a heterogeneous graph do not represent ``similar'' semantics.
Different types of links represent different semantics that need to be specifically dealt with.
Current meta-path-dependent \cite{23_wang2019heterogeneous,27_fu2020magnn} or edge-relationship-oriented \cite{1_zhou2023slotgat,2_lv2021we,3_zhao2022space4hgnn,25_zhu2019relation,26_zhang2019heterogeneous,31_schlichtkrull2018modeling,48_du2023seq} models prone to misconnections in links when there are errors in the graph structure, leading to effectiveness degradation.
Thus, maintaining representation models' effectiveness in noised heterogeneous graphs is still challenging.

To address this challenge, we propose a novel NoiseHGNN model compatible with noised heterogeneous graph learning.
Since the similarity and heterogeneous graphs represent different semantics, we do not calculate their mutual information to correct the noised graph.
Instead, two novel modules, Similarity-aware HGNN, and Metapath-Target contrastive learning, are proposed.
In the Similarity-aware HGNN module, the synthesized similarity graph is used to reinforce the node attention during the representation learning instead of directly supervising the noised graph.
In the Metapath-Target contrastive module, instead of directly contrasting synthesized and noised graphs, we contrast the metapath-based graph (extracted from the noised graph) and the target-based graph (extracted from the synthesized graph) that represent the same semantics.
Then, the representations of both the noised and synthesized graphs are jointly utilized to predict the labels during model training. Finally, the noised graph representation is used to predict the label during the testing.
Extensive experiments on five real-world datasets demonstrate the effectiveness and generalization ability of NoiseHGNN. 
The proposed NoiseHGNN achieves state-of-the-art records on four out of five extensively benchmark datasets under the noised data environment.
In highlights, in the complex and noise-sensitive dataset (DBLP, PubMed, and IMDB), +3 $\sim$ +6\% improvement is observed compared with peer methods. In the rest of the datasets, +1$\sim$ +2\% improvement is observed compared with peer methods.


In summary, our contributions are as follows:

\begin{itemize}
    \item It is the first work, to our best knowledge, to investigate noised heterogeneous graph representation learning and achieves state-of-the-art records.
    \item The proposed Metapath-Target contrastive learning bridges the homogeneity assumption to the noised heterogeneous graph representation.
    \item The proposed similarity-aware HGNN takes advantage of the similarity graph from traditional homogeneous graphs to the noised heterogeneous graph representation learning.
\end{itemize}

    
    




%% file: section/03_Related_Work.tex
\begin{figure*}
    \centering
    \includegraphics[width=0.98\linewidth]{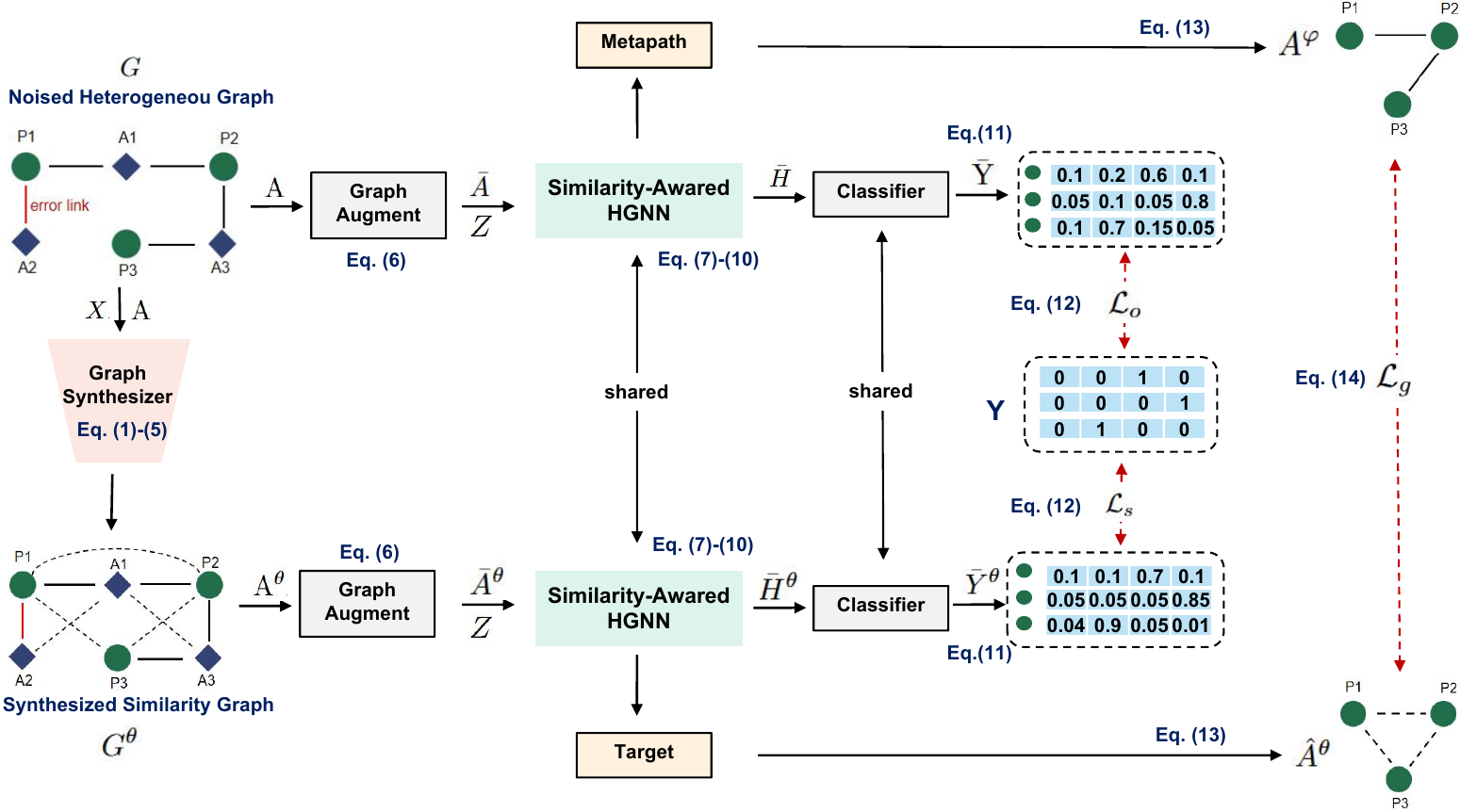}
    \caption{The overall framework of the proposed model.}
    \label{fig: model}
\end{figure*}

\section{Related Work}

\subsection{Graph Structure Learning}
For homogeneous graphs, existing methods address erroneous link perturbations by constructing and refining the graph structure based on the homogeneous graph assumption through graph structure learning. These methods correct the structure of noisy graphs by synthesizing similarity graphs using features of the original nodes. Specifically, existing methods employ probabilistic models \cite{30_wang2021graph,40_franceschi2019learning} and metric learning models \cite{39_fatemi2021slaps,40_franceschi2019learning,41_jin2020graph,42_liu2022towards} to parameterize the adjacency matrix and jointly optimize the parameters of the adjacency matrix and GNNs by solving downstream models. However, in heterogeneous graphs, similar nodes are often not directly connected, making it difficult to use these methods to correct the original noisy graph.
For heterogeneous graphs, HGSL \cite{zhao2021heterogeneous} simultaneously performs Heterogeneous Graph Structure Learning and GNN parameter learning for classification. However, it does not account for the potential noise introduced by erroneous edges within the heterogeneous graph structure.





\subsection{Heterogeneous Graph Neural Network}


HGNNs can generally be categorized into two types based on their strategies for handling heterogeneity:
message-passing based HGNNs, meta-paths based HGNNs.


\textbf{Message-passing based HGNNs.}
RGCN \cite{31_schlichtkrull2018modeling} assigns different weight matrices to various relation types and aggregates one-hop neighbors. RSHN\cite{25_zhu2019relation} builds a coarsened line graph to get edge features and adopts message passing to propagate node and edge features. 
SimpleHGN\cite{2_lv2021we} incorporates relational weight matrices and embeddings to characterize heterogeneous attention at each edge. 
Additionally, Space4HGNN \cite{3_zhao2022space4hgnn} defines a unified design space for HGNNs to exhaustively evaluate combinations of multiple technologies. 
SlotGAT \cite{1_zhou2023slotgat} designs a slot for each type of node according to the node type and uses the slot attention mechanism to construct the HGNN model. 

\textbf{Meta-paths based HGNNs.} 
Another class of HGNNs captures higher-order semantic information through meta-paths.
HAN\cite{23_wang2019heterogeneous} employs hierarchical attention mechanisms to capture both node-level importance between nodes and the semantic-level importance of meta-paths. MAGNN\cite{27_fu2020magnn} enhances this approach with several meta-path encoders to comprehensively encode information along meta-paths. In contrast, the Graph Transformation Network\cite{24_yun2019graph} (GTN) can automatically learn meta-paths through graph transformation layers. However, for heterogeneous graphs with multiple edge types, meta-path-based methods are less practical due to the high cost of acquiring meta-paths. HetGNN \cite{26_zhang2019heterogeneous} addresses this issue by using random walks to sample fixed-size neighbors for nodes of different types and then applying recurrent neural networks (RNNs) for representation learning. 
Seq-HGNN \cite{48_du2023seq} designs a sequential node representation learning mechanism to represent each node as a sequence of meta-path representations during the node message passing.

%% file: section/02_Preliminaries.tex
\section{Preliminaries} 
\subsection{Heterogeneous Graph}
A heterogeneous graph \cite{32_sun2012mining} can be defined as $G=\{\mathcal{V}, \mathcal{E},\phi,\psi\}$, where $\mathcal{V}$ is the set of nodes and $\mathcal{E}$ is the set of edges. Each node  $v$ has a type $\phi(v)$ and each edge $e$ has a type $\psi(e)$. The sets of possible node types and edge types are denoted by $T_v=\{\phi(v):\forall v\in \mathcal{V}\}$ and $T_e=\{\psi(e):\forall e\in \mathcal{E}\}$, respectively. For a heterogeneous graph $|\psi|+|\phi|>2$. When $|T_v|=|T_e|=1$, the graph degenerates into an ordinary homogeneous graph. 

A node $v$ has a feature vector $\mathbf{x}_{v}$. For node type $t\in\Phi $, all type-$t$ nodes $v\in\{v\in\mathcal{V}|\phi(v)=t\}$ have the same feature dimension $d_0^t=d_0^{\phi(v)}$,i.e., $\mathrm{x}_v\in\mathbb{R}^{d_0^{\phi(v)}}$. Nodes of different types can have different feature dimensions  \cite{2_lv2021we}. For input nodes feature type, we use $\eta = 0$ to denote using all given features, $\eta = 1$ to denote using only target node features, and $\eta = 2$ to denote all nodes with one-hot features.

\subsection{Noised Heterogeneous Graph}
Let $G=\{\mathcal{V}, \mathcal{E},\phi,\psi\}$ denote a heterogeneous graph. $N$ be the number of nodes, i.e., $N=|\mathcal{V}|$, and $M$ be the number of links, i.e., $M=|\mathcal{E}|$. The sets of node types is denoted by $T_v=\{\phi(v):\forall v\in \mathcal{V}\}$. An edge $e$ corresponds to an edge type $\psi(e)$ connecting two types of nodes $v_i$ and $v_j$, with the node types  $\phi(v_i)$ and $\phi(v_j)$ respectively.

We simulate the erroneous link scenario in real data by modifying the target node $v_j$ from $T_{v_j}$ connected to $v_i$. Specifically, for all datasets, we randomly modify the target nodes of $30\%$ of the $M$ links, preserving the heterogeneous graph edge type $\psi(e)$.  This ensures that the modified edge types remain consistent with their original types.

%% file: section/04_Proposed_Method.tex
\section{The Proposed Method}



The proposed model mainly consists of four modules: a graph synthesizer, a graph augmenter, a similarity-aware graph encoder, and a graph contrastive module, as shown in Fig \ref{fig: model}.
The synthesized similarity graph contains high-order semantics used to adapt the noise link weights by the similarity-aware graph encoder.
The contrastive learning between the target graph (from the synthesized graph) and the meta-path graph (from the noised graph) can further alleviate noise links.

\subsection{Similarity Graph Synthesizer}


In this work, the synthesized graph is a similarity graph.
The base assumption of the similarity graph is that if two nodes are similar, they will likely have natural relationships. Thus, the similarity graph can potentially supervise the noise graph.
Two processes are essential to synthesize a similarity graph: (1) node feature projection and (2) similarity adjacent matrix synthesizing.

\textbf{Node Feature Projection.} The node features need to be projected into a unified feature space due to the heterogeneity of nodes.
Therefore, we design type-specific linear transformations to project the features of different types of nodes $X^{\phi}$ into the unified feature space $Z$, as defined in equation \ref{eq:projection}.


\begin{equation}
\begin{aligned}
    Z &=\begin{bmatrix}
    Z^{\phi_1}=W^{\phi_1}X^{\phi_1}+b^{\phi_1}\\ Z^{\phi_2}=W^{\phi_2}X^{\phi_2}+b^{\phi_2}\\  \vdots \\ Z^{\phi_{m}}=W^{\phi_{m}}X^{\phi_{m}}+b^{\phi_{m}}
    \end{bmatrix} = \begin{bmatrix} z_1\\z_2 \\ \vdots \\z_{N}\end{bmatrix}
\end{aligned}
\label{eq:projection}
\end{equation}

where $\phi_{m}$ denotes the node type, $X^{\phi_{m}}$ represents the feature matrix of the node type $\phi_{m}$, $W^{\phi_{m}}$ is a learnable matrix, $b^{\phi_{m}}$ denotes vector bias. We transform the node features using type-specific linear transformations for nodes with features and use $X^{\phi_{m}}\in\mathbb{R}^{N\times d^{\phi_{m}}_0}$ to denote the node feature matrix and $\mathrm{Z}\in\mathbb{R}^{N\times d'}$ denotes the transformed node features.

\textbf{Graph Feature Projection.}  The graph projection projects the graph-independent
feature $Z$ into the graph-dependent feature $\bar{Z}$.
Let $\eta$ as an indicator to categorize the graph data environment on feature types.
$\eta=0$ indicates that all attributes of nodes are embedded into the node features (e.g., title, description, abstract, and name of a node).
$\eta=1$ means only the target node's attributes are embedded into the node features.
$\eta=2$ denotes the node's one-hot attributes are used.
According to the value of $\eta$, the graph feature projection is defined as equation \ref{eq:projection2}.



\begin{equation}
\bar{Z} = \begin{cases}
    \sigma(ZW + b),  &\eta\in [0]\\
    \\
    \sigma\left(\widetilde{\mathrm{D}}^{-\frac{1}{2}}\widetilde{\mathrm{A}}\widetilde{\mathrm{D}}^{-\frac{1}{2}}ZW\right) , & \eta \in [1,2]
\end{cases} , \bar{Z} =  \left[\begin{matrix}\bar{z}_{1}\\\vdots\\ \bar{z}_{N}\end{matrix}\right]\\
\label{eq:projection2}
\end{equation}

where $ W$ is the parameter matrix, $\sigma$ is a non-linear function that makes training more stable, ${\widetilde{\mathbf{A}}}=\mathbf{A}+\mathbf{I}$ is the adjacency matrix with self-loop while $\widetilde{\mathrm{D}}$ is the degree matrix of $\widetilde{\mathbf{A}}$.

When $\eta=0$, a simple linear projection is applied.
This is because, in this case, the node features already contain the first-order graph features (i.e., attributes of the node).
When $\eta=1,2$, a typical graph convolutional network is used to aggregate topology information into the node feature.
This is because, in this case, the node itself contains no semantics that need to be enriched from the graph structure.

\textbf{Similarity Adjacent Matrix Synthesizing.} 
To construct the adjacent matrix, we first calculate the similarity value node-to-node based on the graph-dependent node feature $\bar{Z}$.
The equation \ref{eq:similarity} presents the process for calculating the node similarity.


\begin{equation}
S_{i,j} = \frac{\bar{z}_i \cdot \bar{z}_j}{ |\bar{z}_i| \cdot |\bar{z}_j|}
\label{eq:similarity}
\end{equation}

where $S_{i,j}$ is the cosine similarity between feature $\bar{z_i}$ and $\bar{z_j}$.
$\mathrm{S} \in \mathbb{R}^{N \times N}$, and $N$ is the total number of graph node.





Similarity adjacent matrix $\mathrm{S}$ is usually dense and represents fully connected graph structures, which are often not meaningful for most applications and can lead to expensive computational costs \cite{30_wang2021graph}. Therefore, we apply the k-nearest neighbors (kNN)-based sparsification on $\mathrm{S}$. 
Specifically, we retain the links with the top-k connection values for each node and set the rest to zero. 
Let $A^\theta$ represent the Similarity adjacent matrix, as defined in equation \ref{eq:adjacent}.

\begin{equation}
{\mathrm{A}_{ij}^{\theta}}=\begin{cases}{\mathrm{S}}_{ij},&\quad{\mathrm{S}}_{ij}\in\text{top-k}({\mathrm{S}}_{i}),\\0,&\quad{\mathrm{S}}_{ij}\notin\text{top-k}({\mathrm{S}}_{i}),
\end{cases}
\label{eq:adjacent}
\end{equation}
where $\text{top-k}({\mathrm{S_i}}$) is the set of top-k values of row vector ${\mathrm{S_i}}$. For large-scale graphs, we perform the kNN sparsification with its locality-sensitive approximation  \cite{39_fatemi2021slaps} where the nearest neighbors are selected from a batch of nodes instead of all nodes, reducing the memory requirement.

At last, combining the graph-independent node feature $Z$ and the synthesized adjacent matrix $A^\theta$, the synthesized similarity graph $G^\theta$ can be expressed as follows:
\begin{equation}
G^{\theta} = \{Z, A^{\theta}\}
\label{eq: synthetic graph}
\end{equation}

\subsection{Graph Augmentation}

Augmentation is widely used in graph contrastive learning and representation learning.
It can enhance mutual information and improve the model's generalization ability.
In this work, we apply the masking mechanism to augment the graph.
In detail, for a given adjacency matrix $\mathrm{A}$, we first sample a masking matrix $\mathbf{M} \in \{0,1\}^{N \times N}$, where each element of $M$ is drawn independently from a Bernoulli distribution with probability $p^{(A)}$. In NoiseHGNN, we use this graph enhancement scheme to generate enhanced graphs from both the noise and synthesized graphs. The adjacency matrix is then masked with $M$ and $M^{\theta}$:

\begin{equation}
\begin{aligned}
    \bar{A} =& A \odot M \\
    \bar{A}^{\theta} =& A^{\theta} \odot M^{\theta} \\
\end{aligned}
\label{eq: graph augment}
\end{equation}

where  $\bar{\mathrm{A}}$ and $\bar{A}^{\theta} $ are the augmented noise graph and augmented synthesized graph, respectively. To obtain different context structures in the two views, edge discarding for the two views is performed with different probabilities $p^{(A)} \neq p^{(A^{\theta})}$. Other advanced enhancement schemes can also be applied to NoiseHGNN, which is left for future research.

\subsection{Similarity-Aware HGNN Encoder}
The base idea of the similarity-aware encoder is to aggregate the neighbor feature through attention and similarity.
In detail, the primary process of the similarity-aware encoder is (1) correlation coefficient, (2) similarity-aware attention, and (3) attention-based aggregation.

\textbf{Correlation coefficient.}  The same as the classic Graph Attention Network (GAT), we first calculate the correlation coefficient $e_{i,j}$ between a node $i$ and its neighbors $j$,  as the equation \ref{eq:correlation} shows.

\begin{equation}
\begin{aligned}
    e_{i,j} = a ([Wz_i || Wz_j ]), j \in \mathtt{Index}(\bar{A}_i)\\
    e^{\theta}_{i,j} =  a ([Wz_i || Wz_j ]), j \in \mathtt{Index}( \bar{A}^{\theta}_i )\\
\end{aligned}
\label{eq:correlation}
\end{equation}

Here, $W$ is a learnable parameter, while $z$ is the projected node feature. 
$a(\cdot)$ is a projection network that projects concatenated features to a scalar number ranging from 0 to 1.0. 
$\mathtt{Index}(\bar{A}_i)$ denotes a set that contains all indexes of nodes that are neighbors of node $i$ in the noise graph.

$\mathtt{Index}(\bar{A}^\theta_i)$ denotes a set that contains all indexes of nodes that are neighbors of node $i$ in the synthesized similarity graph $G^\theta$.

\textbf{Similarity-aware attention.} Similar to GAT, we calculate the attention $a_{i,j}$ through the $\mathtt{Softmax(\cdot)}$ operation on correlation coefficient $e_{i,j}$.
However, the difference is that the proposed attention also considers the similarity graph's adjacent matrix, as shown in equation \ref{eq:attention}.
\begin{equation}
\begin{aligned}
    \alpha_{i,j} = \frac{\mathtt{exp}(\mathtt{LeakyReLU}(e_{i,j}))}{\sum_{k \in \mathtt{Index}(\bar{A}_i)}\mathtt{exp}(\mathtt{LeakyReLU}(e_{i,k}))} \cdot \bar{A}_{i,j}\\
    \alpha^{\theta}_{i,j} =  \frac{\mathtt{exp}(\mathtt{LeakyReLU}(e^{\theta}_{i,j}))}{\sum_{k \in \mathtt{Index}(\bar{A}^{\theta}_i)}\mathtt{exp}(\mathtt{LeakyReLU}(e^{\theta}_{i,k}))}  \cdot  \bar{A}^{\theta}_{i,j} \\
\end{aligned}
\label{eq:attention}
\end{equation}

From the above attention, the similar node pairs will have a higher attention coefficient, while the dissimilar node pairs will only have zero attention value.
This potentially intercepts the message propagation on error-prone links when the corresponding attention is zero.

\textbf{Attention-based aggregation.} The aggregation aims to obtain the graph representation $H$ from node feature $Z$ and adjacent matrix $A$.
Let $W^{\alpha}$ denote a learnable parameter of the aggregation process.
The noise and synthesized similarity graphs share the same $W^{\alpha}$.
$\sigma(\cdot)$ denote a $\mathtt{LeakyReLU(\cdot)}$ operation.
Equation \ref{eq:representations} defines the node representation calculated by similarity-aware attention $\alpha$ and node feature $Z$.



\begin{equation}
\begin{aligned}
\bar{h}_i=\sigma \Bigg( \sum_{j \in {\mathtt{Index}(\bar{A}_i})} \alpha_{ij} \cdot W^{\alpha} \cdot {z}_j \Bigg) \\
\bar{h}^{\theta}_i=\sigma \Bigg( \sum_{j \in {\mathtt{Index}(\bar{A}^{\theta}_i)}} \alpha^{\theta}_{ij} \cdot W^{\alpha} \cdot {z}_j \Bigg) 
\end{aligned}
\label{eq:representations}
\end{equation}

where $j \in {\mathtt{Index}(\bar{A}_i)} $ and $j \in {\mathtt{Index}(\bar{A}^{\theta}_i)}$ represent the neighbors of node $i$ under the noise and synthesized graphs respectively.
$\bar{h}_i$ and $\bar{h}^{\theta}_i$ represent the node representations of node $v_i$ under the noise and synthesized similarity graphs, respectively.

Finally, all nodes get the graph representation $H$, which can be represented as follows:
\begin{equation}
\begin{aligned}
\bar{H}= \left[\begin{matrix}h_{1}\\\vdots\\h_{N}\end{matrix}\right] , 
\bar{H}^{\theta} =  \left[\begin{matrix}h^{\theta}_{1}\\\vdots\\h^{\theta}_{N}\end{matrix}\right]
\end{aligned}
\label{eq: all node representations}
\end{equation}
where $\bar{H}$ denotes all node representations under the noise graph, while $\bar{H}^{\theta}$ denotes all node representations under the synthesized similarity graph.

\subsection{Graph Learning}
\textbf{Learning Objective.} 
So far, we have the graph representations $\bar{H}$ and $\bar{H}^\theta$.
The representations can be used for multiple downstream objectives such as node classifications, graph classifications, clustering, link prediction, etc.
This work focuses on node classifications, which categorize a node instance into a class label.
Let $\mathtt{MLP(\cdot)}$ represent a linear neural network. 
The learning objective ( node classification) is defined as follows:

\begin{equation}
\begin{aligned}
\bar{{\mathrm{Y}}} &= \mathtt{Softmax}(\mathtt{MLP}(\bar{H})) ,\\
\bar{Y}^{\theta}  &= \mathtt{Softmax}(\mathtt{MLP}(\bar{H}^{\theta})) 
\end{aligned}
\label{eq:objective}
\end{equation}
where $\bar{{\mathrm{Y}}}$ and $\bar{Y}^{\theta}$ are the predictions obtained under the noise graph and the synthesized similarity graph, respectively.

\textbf{Classification loss.} 
The datasets normally have two characteristics: single-label and multi-label.
For single-label classification,  softmax and cross-entropy loss are used. 
For multi-label datasets, sigmoid activation and binary cross-entropy loss are used. 
Let $\mathcal{L}_{CE}(\cdot)$ represent the cross-entropy for both single-label and multi-label in general.
$Y$ is the ground truth label of the classification.
The classification loss is defined as equation \ref{eq: CE_LOSS}.

\begin{equation}
\begin{aligned}
\mathcal{L}_o =& \mathcal{L}_{CE}(\bar{{Y}},\mathrm{Y})  \\
\mathcal{L}_s =& \mathcal{L}_{CE}(\bar{Y}^{\theta},\mathrm{Y})
\end{aligned}
\label{eq: CE_LOSS}
\end{equation}
where $\mathcal{L}_{o}$ is the classification loss for the noise graph representation.
$\mathcal{L}_{s}$ is the classification loss for the synthesized similarity graph representation.

\textbf{Contrastive Loss.} To better mitigate the error links in the noise graph structure, we extract the meta-path $A^{\varphi}$ graph from the noise graph structure (e.g., author-paper-author (APA) in the DBLP) and also extract the target graph $\hat{A}^{\theta}$ from the synthesized similarity graph, as presented in equation \ref{eq: subgraph}.

\begin{equation}
    \begin{aligned}
    A^{\varphi} = &\mathtt{MetaPath}(A) \\
    \hat{A}^{\theta} = &\mathtt{Target}({A}^{\theta})
    \end{aligned} 
    \label{eq: subgraph}
\end{equation}


Based on the above subgraphs, we then use the scaled cosine loss function to calculate graph contrastive loss $\mathcal{L}_{g}$. 
Since the higher-order synthesized similarity graph is an adjacency matrix with higher-order semantic information, it can alleviate the errors in the meta-paths graph. 
Specifically, the contrastive loss $\mathcal{L}_{g}$ is calculated as follows:

\begin{equation}
\mathcal{L}_{g}=(1-\frac{(A^{\varphi})^{T} \cdot \hat{A}^{\theta}}{\|(A^{\varphi})^{T}\|\cdot\|\hat{A}^{\theta}\|})^{\gamma}  , \gamma\geq1
\label{eq: Lg_LOSS}
\end{equation}
where the scaling factor ${\gamma}$ is a hyper-parameter adjustable over different datasets, $A^{\varphi}$ denotes the graph obtained from the noise graph via meta-paths and $\hat{A}^{\theta}$ denotes the target graph extracted from the synthesized graph.

\textbf{Final Loss.} Combining equation \ref{eq: CE_LOSS} and \ref{eq: subgraph}, the final loss function is formulated as:

\begin{equation}
\mathcal{L} = \mathcal{L}_{o} + \mathcal{L}_{s} + \mathcal{L}_{g}
\label{eq: final objective function}
\end{equation}

The whole process of the above equations is organized as a pseudo-code algorithm, see the appendix for details.




%% file: section/05_Experiments.tex
\begin{table*}[!ht]
\centering
\resizebox{\textwidth}{!}{%
\begin{tabular}{ccccccccccc}
\toprule
\textbf{} & \multicolumn{2}{c}{\textbf{DBLP(0.3)}}    & \multicolumn{2}{c}{\textbf{IMDB(0.3)}}    & \multicolumn{2}{c}{\textbf{ACM(0.3)}}     & \multicolumn{2}{c}{\textbf{PubMed\_NC(0.3)}} & \multicolumn{2}{c}{\textbf{Freebase(0.3)}} \\
\midrule
          & Macro-Fl            & Micro-Fl            & Macro-F1            & Micro-F1            & Macro-Fl            & Micro-Fl            & Macro-Fl              & Micro-F1             & Macro-F1             & Micro-F1            \\
\midrule
GCN (2016)       & 50.97±0.14          & 52.35±0.17          & 37.86±3.93          & 50.60±1.73          & 86.38±1.21          & 86.46±1.14          & 36.28±2.78            & 41.39±2.81           & 17.16±0.22           & 53.66±0.18          \\

GAT (2017)      & 63.11±1.48          & 64.38±1.23          & 38.53±5.65          & 49.65±4.58          & 79.02±3.35          & 79.61.60±3.26       & 34.08±4.66            & 39.53±3.89           & 17.83±0.73           & 54.63±0.23          \\
RGCN (2018)     & 43.02±1.83          & 45.02±1.53          & 39.79±1.77          & 47.28±0.90          & 54.37±2.38          & 55.36±2.25          & 15.28±3.30            & 19.76±3.96           & -                    & -                   \\
RSHN (2019)     & 67.01±1.47          & 67.69±1.39          & 31.40±3.92          & 48.09±1.45          & 80.43±2.09          & 80.43±1.83          & -                     & -                    & -                    & -                   \\
HAN (2019)      & 62.32±1.79          & 63.43±1.59          & 38.64±4.7           & 49.93±2.97          & 78.31±1.29          & 78.19±1.37          & -                     & -                    & -            & -          \\
HetGNN (2019)    & 59.31±0.23          & 60.23±0.36          & 37.36±0.75          & 40.85±0.86          & 74.45±0.13          & 74.39±0.15          & -                     & -                    & -                    & -                   \\
MAGNN (2020)    & 60.83±1.40          & 62.66±1.07          & 14.30±0.20          & 39.98±0.12          & 88.67±0.84          & 88.64±0.73          & -                     & -                    & -                    & -                   \\
HGT (2020)       & 48.11±4.39          & 53.66±2.54          & 45.26±0.89          & 55.62±0.16          & 81.52±1.82          & 81.43±1.77          & \underline {46.65±3.63 }           & 49.53±2.61           & 10.40±0.94           & 49.31±0.13          \\
simpleHGN (2021) & 65.54±1.03          & 67.63±0.75          & \underline{58.21±2.20}    & \underline{62.93±0.92}    & 89.65±0.52          & 89.64±0.48          & 42.25±4.67            & 47.90±3.40           & 27.10±1.54           & \underline{56.79±0.48}    \\
space4HGNN(2022)       & 65.43±1.12         & 67.81±0.55       & 55.43±0.54          & 61.91±0.94          & 88.52±0.74          & 88.63±0.34          & 42.41±3.59            & 47.14±3.16           & 26.93±0.78           & 53.54±0.33          \\
Seq-HGNN(2023)       & 67.89±0.26         & 68.93±0.37       & 56.23±0.43          & 62.34±0.29          & 90.02±0.23          & 90.09±0.31          & 44.41±4.31            & 48.54±4.01           & 27.93±0.84           & 55.94±0.48        \\
SlotGAT(2024)  & \underline{68.23±0.75}    & \underline{69.77±0.19}    & 54.89±1.77          & 62.47±0.97          & \textbf{90.64±0.46} & \textbf{90.65±0.42} & { 44.76±3.36}      & \underline{50.88±1.54}     & \underline{28.14±1.31}     & 56.56±0.64          \\
\midrule
NoiseHGNN      & \textbf{71.96±0.33} & \textbf{73.16±0.19} & \textbf{60.46±0.69} & \textbf{63.94±0.26} & \underline{90.16±0.49}    & \underline{90.10±0.46}    & \textbf{50.92±3.52}   & \textbf{55.81±3.20}  & \textbf{30.57±2.94}  & \textbf{56.96±0.67}
 \\
\bottomrule
\end{tabular}
}
\caption{Peer comparison on node classification task. 30\% error link rate is applied to the datasets.  Vacant positions (“-”) mean out of memory in our computational environment. The best record is marked in bold, and the runner-up is underlined.}
\label{tab:my-table2}
\end{table*}

\section{Experiments}
\subsection{Experiment Settings}

\textbf{Datasets.} 
Table \ref{tab: table_dataset} reports the statistics of the benchmark datasets widely used in previous studies  \cite{2_lv2021we,1_zhou2023slotgat,3_zhao2022space4hgnn}. These datasets span various domains, such as academic graphs (e.g., DBLP, ACM), information graphs (e.g., IMDB, Freebase), and medical-biological graphs (e.g., PubMed). For node classification, each dataset contains a target node type, and all nodes of this target type are used for classification.

\begin{table}[!ht]
\centering
\resizebox{\columnwidth}{!}{%
\begin{tabular}{cccccccc}
\toprule
\begin{tabular}[c]{@{}c@{}}Node\\ Classification\end{tabular} & Nodes & \begin{tabular}[c]{@{}c@{}}Node\\ Types\end{tabular} & Edges   & \begin{tabular}[c]{@{}c@{}}Edge\\ Types\end{tabular} & \begin{tabular}[c]{@{}c@{}} error\\ link\end{tabular} & Target  & Classes \\
\midrule
DBLP                                                          & 26,128  & 4                                                      & 239,566   & 6                                                      & 30\%                                                    & author  & 4         \\
IMDB                                                          & 21.420  & 4                                                      & 86.642    & 6                                                      & 30\%                                                    & movie   & 5         \\
ACM                                                           & 10,942  & 4                                                      & 547,872   & 8                                                      & 30\%                                                    & paper   & 3         \\
Freebase                                                      & 180,098 & 8                                                      & 1,057,688 & 36                                                     & 30\%                                                    & book    & 7         \\
PubMed\_NC                                                    & 63,109  & 4                                                      & 244,986   & 10                                                     & 30\%                                                    & disease & 8           \\
\bottomrule
\end{tabular}%
}
\caption{Statistics of Datasets.}
\label{tab: table_dataset}
\end{table}

\textbf{Baselines.} We compare our method with several state-of-the-art models, including HAN  \cite{23_wang2019heterogeneous}, MAGNN  \cite{27_fu2020magnn}, HetGNN \cite{26_zhang2019heterogeneous}, HGT  \cite{28_hu2020heterogeneous}, RGCN \cite{31_schlichtkrull2018modeling}, RSHN \cite{25_zhu2019relation}, SimpleHGN \cite{2_lv2021we}, Space4HGNN \cite{3_zhao2022space4hgnn}, Seq-HGNN\cite{48_du2023seq} and SlotGAT \cite{1_zhou2023slotgat}. Additionally, we include comparisons with GCN \cite{35_kipf2016gcn} and GAT \cite{36_velivckovic2017gat}.




\textbf{Evaluation Settings.} For node classification, following the methodology described by Lv et al. (2021), we split the labeled training set into training and validation subsets at 80\%:20\%. The testing data are fixed, with detailed numbers provided in the appendix. For each dataset, we conduct experiments on five random splits, reporting the average and standard deviation of the results. For our methods, we perform a grid search to select the best hyperparameters on the validation set. The results of the node classification are reported using the averaged Macro-F1 and Micro-F1 scores.


\subsection{Main Results}

Table \ref{tab:my-table2} presents the node classification results on a 30\% noise data environment compared with peer methods.
The comparison metrics are mean Macro-F1 and Micro-F1 scores.
It can be observed that NoiseHGNN consistently outperforms all baselines for Macro-F1 and Micro-F1 on DBLP, IMDB, Pubmed, and Freebase. 
Overall, NoiseHGNN achieves an average +4.2\% improvement on Macro-F1 and +2.6\% on Micro-F1 compared with runner-up.
In highlight, NoiseHGNN achieves the best results on IMDB and Pubmed datasets, with an average improvement of +5.9\%.
This is because IMDB and Pubmed are noise-sensitive datasets since their node features are just one-hot embedding, which contains no semantics.
All the information is stored in the topology of the graphs, which have 30\% error links.
In the medium-size dataset, DBLP, NoiseHGNN performs stable, with a +3.5\% improvement compared with the previous SOTA method.
In the large and non-trivial dataset, Freebase, NoiseHGNN still obtained 30.57\%  Macro-F1, which is 2.4\% higher than the previous SOTA method.
Notably, NoiseHGNN gets the runner-up results (90.16\%) in the ACM dataset, with -0.48\% lower than SlotGAT (90.64\%).
This is because ACM is a noise-insensitive and trivial dataset, while the latest methods can achieve 90\% and even higher results.
Its nodes contain rich semantic embedding that can be simply used for node classification without using graph links.



\subsection{Ablation Study}
In this section, we demonstrate the effectiveness of NoiseHGNN through ablation experiments with w/o Graph Synthesized and Meta-Target Graph. As shown in table\ref{tab: ablation_graph_synthesizer}, NoiseHGNN with graph synthesized and Meta-target graph achieves superior results across all datasets. 
Specifically, on the smaller DBLP dataset, NoiseHGNN with the graph synthesizer and Meta-target graph improves Macro-F1 and Micro-F1 scores by 2.41\% and 3.81\%, respectively. On the largest dataset, Freebase, the improvements are 11.40\% in Macro-F1 and 3.24\% in Micro-F1. Notably, on the IMDB dataset, NoiseHGNN achieves significant improvements of 42.01\% in Macro-F1 and 25.83\% in Micro-F1. These results demonstrate that the higher-order synthesized graphs generated by the graph synthesizer effectively mitigate the negative effects of erroneous links, leading to substantially better performance.

\begin{table}[htbp!]
\centering
\resizebox{\columnwidth}{!}{%
\begin{tabular}{ccccc}
\toprule
Datasets &  & w/o Graph Synthesizer & w/o Meta-Target Graph & w/ All \\ 
\midrule
\multirow{2}{*}{\textbf{DBLP}} & Macro-Fl & 70.26±1.23 & 71.66±0.58 & \textbf{71.96±0.33} \\
 & Micro-Fl & 70.47±1.28 & 72.78±0.58 & \textbf{73.16±0.19} \\ \midrule
\multirow{2}{*}{IMDB} & Macro-Fl & 43.46±6.58 & 59.77±1.77 & \textbf{60.46±0.69} \\
 & Micro-Fl & 50.78±1.29 & 63.46±0.73 & \textbf{63.90±0.26} \\ \midrule
\multirow{2}{*}{ACM} & Macro-Fl & 89.40±0.92 & 89.32±0.61 & \textbf{90.16±0.49} \\
 & Micro-Fl & 89.34±1.09 & 89.42±0.60 & \textbf{90.10±0.46} \\ \midrule
\multirow{2}{*}{Pubmed\_NC} & Macro-Fl & 49.92±1.32 & 49.40±1.74 & \textbf{50.92±3.52} \\
 & Micro-Fl & 51.32±1.52 & 52.42±3.52 & \textbf{55.81±3.20} \\ \midrule
\multirow{2}{*}{Freebase} & Macro-Fl & 27.44±1.96 & 28.16±2.47 & \textbf{30.57±2.94} \\
 & Micro-Fl & 55.14±1.33 & 56.15±0.34 & \textbf{56.96±0.67} \\ \bottomrule
\end{tabular}%
}
\caption{Ablation Study: w/o Graph Synthesized and Meta-Target Graph.}
\label{tab: ablation_graph_synthesizer}
\end{table}

\subsection{Parameter Analysis} 

\begin{figure}[H]
    \centering
    \includegraphics[width=0.95\linewidth]{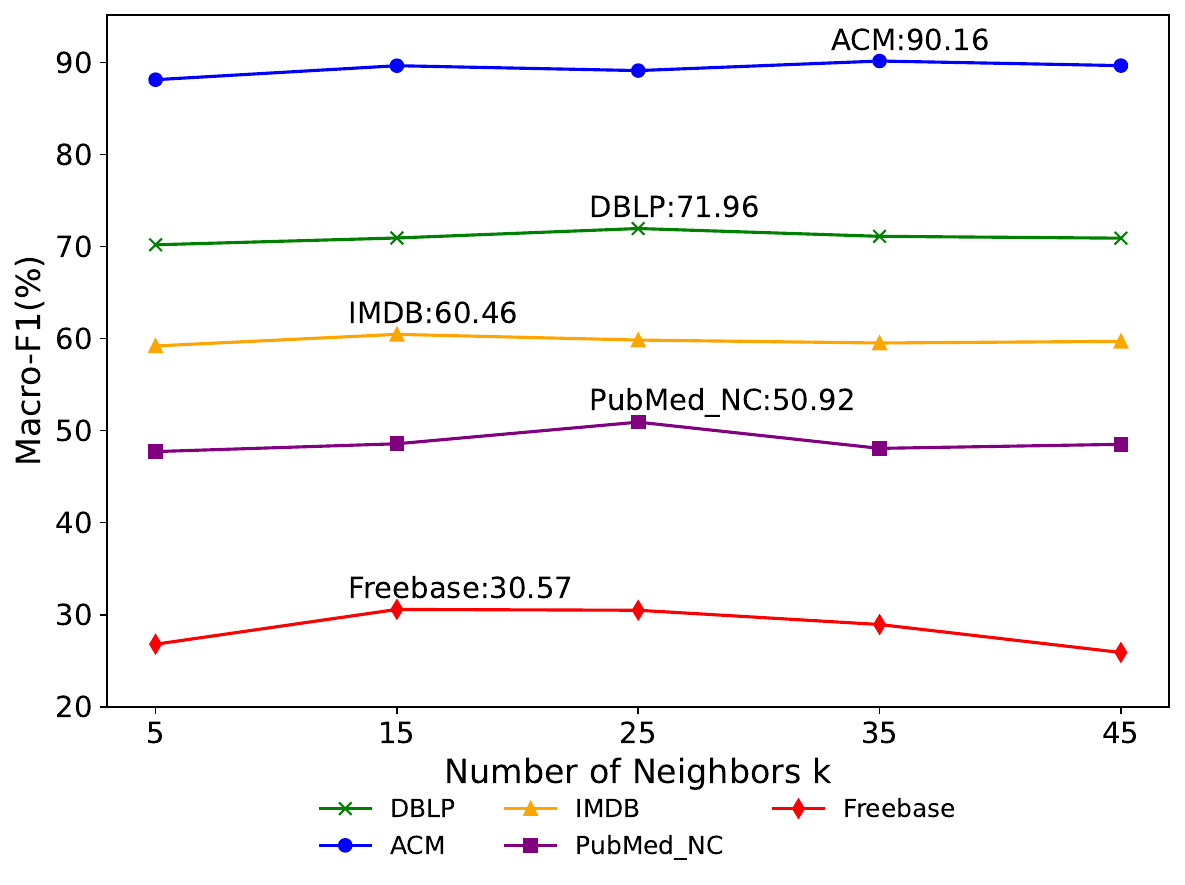}
    \caption{Effect of parameter k in top-k($\cdot$) selection in equation \ref{eq:adjacent}.}
    \label{fig: k-Macro}
\end{figure}

In this subsection, we explore the sensitivity of the hyperparameter $k$ in NoiseHGNN. The parameter $k$ determines the number of neighbor nodes to be used in the S high-order synthesized graph based on k-nearest neighbors (kNN). 
To understand its impact on our model's performance, we search the number of neighbors $k$ in the range of $\{5, 15, 25,35, 45\}$ for all datasets. Our results indicate that selecting an appropriate k value can significantly enhance the accuracy of NoiseHGNN across various datasets. As is demonstrated in Fig. \ref{fig: k-Macro}, the best selection for each dataset is different, i.e., $k=15$ for Freebase and IMDB, $k=25$ for Cora and  PubMed\_NC, and $k=35$ for ACM. It is commonly observed that selecting a value of $k$ that is either too large or too small can lead to suboptimal performance. We hypothesize that an excessively small $k$ may restrict the inclusion of beneficial neighbors, while an overly large $k$ might introduce noisy connections, thereby degrading the overall performance.



\subsection{Visualization Analysis}

\begin{figure} [!htbp]
    \centering
    \begin{minipage} {0.49\linewidth}
        \centering
        \includegraphics[width=1\linewidth]{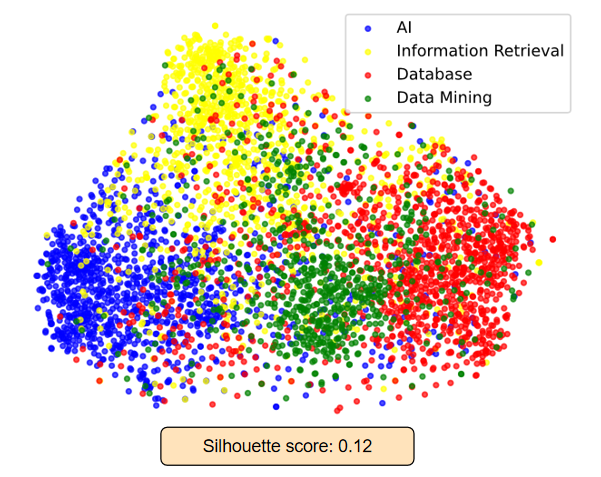}
    \end{minipage}
    \begin{minipage} {0.49\linewidth}
        \centering
        \includegraphics[width=1\linewidth]{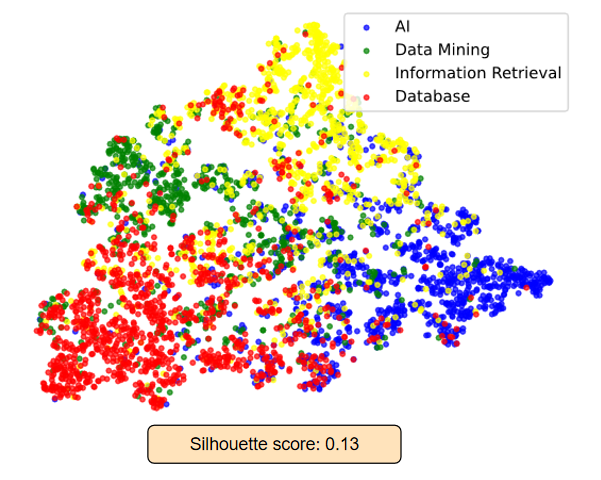}
    \end{minipage}
    \quad
    \begin{minipage} {0.55\linewidth}
        \centering
        \includegraphics[width=1\linewidth]{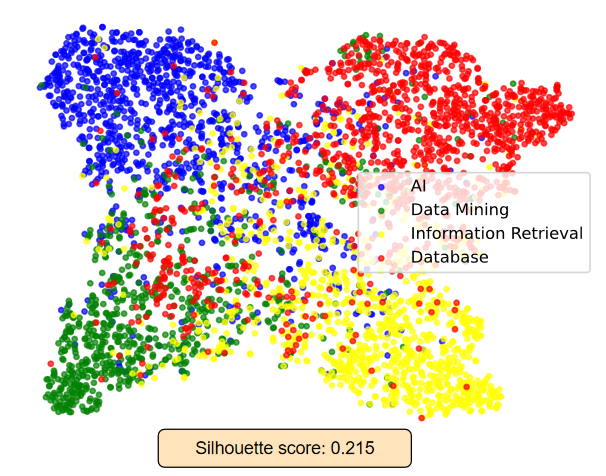}
    \end{minipage}
    \caption{The t-SNE visualization of the graph representation.  The proposed method achieves a good silhouette score (0.215).}
    \label{fig:t-sne}
\end{figure}

To visually represent and compare the quality of the embeddings, figure \ref{fig:t-sne} presents the t-SNE plot  \cite{46_van2008visualizing} of the node embeddings generated by NoiseHGNN on the DBLP dataset. Consistent with the quantitative results, the 2D projections of the embeddings learned by NoiseHGNN show more distinguishable clusters, both visually and numerically, compared to simpleHGN and SlotGAT. Specifically, the category of Data Mining, NoiseHGNN is better able to differentiate itself from other categories, while neither simpleHGN nor SlotGAT can do so well. The Silhouette scores support this \cite{47_rousseeuw1987silhouettes}, where NoiseHGNN achieves a score of 0.215 on DBLP, significantly higher than the scores of 0.12 for simpleHGN and 0.13 for SlotGAT.

%% file: section/06_Conclusion.tex
\section{Conclusion, and Future Work}

This paper presents the first study addressing the problem of error link perturbation in heterogeneous graphs. To tackle this issue, we propose a novel method, \textbf{NoiseHGNN}, which effectively extracts valid structural information from the original data, thereby mitigating the negative impact of erroneous links. Extensive experiments on numerous real-world datasets demonstrate that the proposed method achieves state-of-the-art performance in noisy heterogeneous graph learning tasks.
In highlights, +5$\sim$6\% improvements are observed in several noised datasets compared with previous SOTA methods. 

In the future, we intend to explore the performance of heterogeneous graphs in complex environments. Specifically, we intend to address the problem of model failure in the presence of missing features, missing links, and erroneous link perturbations.
We hope to advance the robustness and applicability of heterogeneous graph neural networks in more complex and uncertain real-world scenarios.

%% file: section/07_appendix.tex

\begin{appendices}

\section{Homogeneity Assumption Analysis}
In a homogeneous graph, the similarity graph is potentially close to the original graph.
However,  the similarity graph has a different topology from the original graph in a heterogeneous data environment.
In this work, we propose a novel way to bridge the similarity graph with the heterogeneous graph using the target and meta-path graphs.
The target graph can extract the homogeneity information from the synthesized graph, while the meta-path graph can be from the original heterogeneous graph.
Here, we try to quantitatively analyze the target graph's homogeneity score compared to the meta-path graph.
Note that the target graph is synthesized without noise links, while the meta-path graph is disturbed by noise links.
This means the higher the score obtained by the target graph, the higher the ability to alleviate the noise disturbance.


Following the previous work\cite{3_zhao2022space4hgnn}, the homogeneity score $\beta$ is defined as follows:

\begin{equation}
\beta = \frac{1}{|V|} \sum_{v \in V}
\frac{| \{ u: \boldsymbol{A}_{\mathcal{P}}[u, v] = 1, y_u = y_v \}|}{| \{ u: \boldsymbol{A}_{\mathcal{P}}[u, v] = 1\}|},
\end{equation}

where $y_u$ and $y_v$ represent the label of node $u$ and $v$, respectively.

\begin{table}[H]
\centering

\resizebox{\columnwidth}{!}{%
\begin{tabular}{c|c|c|c}
\hline
\textbf{Dataset} & \textbf{Meaning} & \textbf{Subgraph} & \textbf{$\beta$} \\ \hline
\multirow{9}{*}{\textbf{ACM}} & \multirow{9}{*}{\begin{tabular}[c]{@{}c@{}}P: paper \\  A: author \\ S: subject \\  c: citation relation \\  r: reference relation\end{tabular}} & $A^{\varphi}_{PrP}$ & 0.4991 \\   
 &  & $A^{\varphi}_{PcP}$ & 0.4927 \\   
 &  & $A^{\varphi}_{PAP}$ & \underline{ 0.6511} \\   
 &  & $A^{\varphi}_{PSP}$ & 0.4572 \\   
 &  & $A^{\varphi}_{PcPAP}$ & 0.5012 \\   
 &  & $A^{\varphi}_{PcPSP}$ & 0.4305 \\   
 &  & $A^{\varphi}_{PrPAP}$ & 0.4841 \\   
 &  & $A^{\varphi}_{PrPSP}$ & 0.4204 \\   
 &  & $\hat{A}^{\theta}$ & \textbf{0.6620} \\ \hline
\multirow{4}{*}{\textbf{DBLP}} & \multirow{4}{*}{\begin{tabular}[c]{@{}c@{}}A: author \\ P: paper\\  T: term \\ V: venue\end{tabular}} & $A^{\varphi}_{APA}$ & \underline{ 0.7564} \\   
 &  & $A^{\varphi}_{APTPA}$ & 0.2876 \\   
 &  & $A^{\varphi}_{APVPA}$ & 0.3896 \\   
 &  & $\hat{A}^{\theta}$ & \textbf{0.8008} \\ \hline
\multirow{6}{*}{\textbf{PubMed\_NC}} & \multirow{6}{*}{\begin{tabular}[c]{@{}c@{}}D: disease\\ G: gene \\  C: chemical \\  S: species\end{tabular}} & $A^{\varphi}_{DD}$ & 0.0169 \\   
 &  & $A^{\varphi}_{DCD}$ & 0.1997 \\   
 &  & $A^{\varphi}_{DDD}$ & 0.1945 \\   
 &  & $A^{\varphi}_{DGD}$ & \underline{ 0.2567} \\   
 &  & $A^{\varphi}_{DSD}$ & 0.2477 \\   
 &  & $\hat{A}^{\theta}$ & \textbf{0.3326} \\ \hline
\multirow{8}{*}{\textbf{Freebase}} & \multirow{8}{*}{\begin{tabular}[c]{@{}c@{}}B: book\\  F: film \\  L: location \\  M: music \\  P: person \\  S: sport \\  O: organization \\  U: business\end{tabular}} & $A^{\varphi}_{BB}$ & 0.1733 \\   
 &  & $A^{\varphi}_{BUB}$ & 0.0889 \\   
 &  & $A^{\varphi}_{BFB}$ & 0.1033 \\   
 &  & $A^{\varphi}_{BLMB}$ & 0.0303 \\   
 &  & $A^{\varphi}_{BOFB}$ & \underline{ 0.3341} \\   
 &  & $A^{\varphi}_{BPB}$ & 0.1928 \\   
 &  & $A^{\varphi}_{BPSB}$ & 0.0603 \\   
 &  & $\hat{A}^{\theta}$ & \textbf{0.9361} \\ \hline
\end{tabular}
}
\caption{Homogeneity score comparison between target and meta-path graphs.}
\label{tab: homophily}
\end{table}



As illustrated in Table \ref{tab: homophily}, the homogeneity scores obtained for our target graph are consistently higher than those obtained from the meta-path graph across all datasets. 
In particular, our homogeneity score on the Freebase dataset is significantly higher than those obtained for the meta-path graph.
This is because the meta-path graph inevitably suffers from noise links, but the target graph does not.

\section{Analysis of Graph  synthesizer} 
\label{ap: synthesizer}
In Table \ref{app: properties},
we summarize the properties of the proposed graph synthesizers, including their memory, parameter, and time complexity. Here, $n$ and $d$ denote the number of nodes and the number of node representation dimensions, respectively, while $k$ denotes the number of kNN neighbors and $L$ represents the number of layers of the synthesizer.

We consider the complexities with locality-sensitive kNN sparsification where the neighbors are selected from a batch of nodes (batch size $=b_1$). We provide our analysis as follows: MLP and GCN synthesizers require larger space and time complexity to consider the correlation between features and original topology. With the effective kNN sparsification, the memory and time complexity are reduced from $\mathcal{O}(n^2)$ to $\mathcal{O}(n)$, which improves the scalability of synthesizers.

\begin{table}[H]
\centering

\resizebox{\columnwidth}{!}{%
\begin{tabular}{cccc}
\hline
\textbf{synthesizer} & \textbf{Memory} & \textbf{Params} & \textbf{Time} \\ \hline
MLP & $\mathcal{O}(ndL +nk)$ & $\mathcal{O}(d^2 L)$ & $\mathcal{O}(nd^2L+ndb_1)$ \\ \hline
GCN & $\mathcal{O}(ndL +nk)$ & $\mathcal{O}(d^2 L)$ & $\mathcal{O}(mdL+ nd^2L+ndb_1)$ \\ \hline
\end{tabular}%
}
\caption{Properties of graph synthesizer}
\label{app: properties}
\end{table}

Considering these properties, we assigned the appropriate synthesiser to each dataset. Specifically, for datasets with node feature type $\eta=0$ (e.g., IMDB), we use the MLP synthesizer to model them due to the richness of their node features.
For datasets with node feature type $\eta=2$ (e.g., DBLP), due to the associated low feature dimensions, the GCN synthesizer can be further considered to leverage additional topological information with original graphs.

\section{Efficiency Analysis} \label{ap: Efficiency}


Tables \ref{tab: traintime} and \ref{tab: infertime} present a comparison of the training time and inference time of NoiseHGNN against three strong baselines: simpleHGN, HGT, and SlotGAT. Table \ref{tab: memory} further reports the peak memory usage of these models.

\begin{table}[H]
\centering

\resizebox{\columnwidth}{!}{%
\begin{tabular}{cccccc}
\toprule
 & DBLP & IMDB & ACM & PubMed\_NC & Freebase \\ \midrule
simpleHGN & 50.11 & 49.42 & 67.10 & 79.66 & 293.14 \\ \midrule
SlotGAT & 171.20 & 160.29 & 137.87 & 362.40 & 361.03 \\ \midrule
HGT & 243.08 & 472.98 & 481.12 & 688.55 & 1330.99 \\ \midrule
NoiseHGNN & 104.30 & 100.45 & 49.84 & 278.89 & 1871.17 \\ 
\bottomrule
\end{tabular}%
}
\caption{Training time per epoch (millisecond)}
\label{tab: traintime}
\end{table}

\begin{table}[H]
\centering
\resizebox{\columnwidth}{!}{%
\begin{tabular}{cccccc}
\toprule
 & DBLP & IMDB & ACM & PubMed\_NC & Freebase \\ \midrule
simpleHGN & 5.38 & 7.42 & 5.48 & 5.69 & 94.34 \\ \midrule
SlotGAT & 86.06 & 87.49 & 47.33 & 197.99 & 191.04 \\ \midrule
HGT & 157.76 & 128.46 & 187.59 & 290.13 & 629.81 \\ \midrule
NoiseHGNN & 10.50 & 14.60 & 9.77 & 21.92 & 99.41 \\ \bottomrule
\end{tabular}%
}
\caption{Inference time (millisecond)}
\label{tab: infertime}
\end{table}

\begin{table}[H]
\centering
\resizebox{\columnwidth}{!}{%
\begin{tabular}{cccccc}
\toprule
 & DBLP & IMDB & ACM & PubMed\_NC & Freebase \\ \midrule
simpleHGN & 3.84 & 3.01 & 6.93 & 5.96 & 7.06 \\ \midrule

SlotGAT & 7.97 & 6.34 & 7.57 & 13.45 & 9.61 \\ \midrule
HGT & 1.24 & 1.08 & 1.81 & 2.54 & 6.29 \\ \midrule
NoiseHGNN & 1.60 & 2.06 & 0.92 & 3.27 & 10.67 \\ 
\bottomrule
\end{tabular}%
}
\caption{Peak GPU memory (GB)}
\label{tab: memory}
\end{table}

As demonstrated, NoiseHGNN exhibits superior performance in terms of training time, inference time, and memory usage. and as mentioned it achieves state-of-the-art results. As previously stated, it achieves state-of-the-art results.


\section{Dataset Details}
\label{sec:appendix}
\subsection{Dataset Descriptions}
For all of the benchmark datasets, one could access them in the online platform HGB\footnote{{\url{https://www.biendata.xyz/hgb/}}}.
\begin{itemize}
    \item \textbf{DBLP} is a bibliography website of computer science. There are 4 node types, including authors, papers, terms, and venues, as well as  $6$ edge types. The edge types include paper-term, paper-term, paper-venue, paper-author, term-paper, and venue-paper. The target is to predict the class labels of authors. The classes are database, data mining, AI, and information retrieval. 

    \item \textbf{ACM.} This heterogeneous academic graph encompasses nodes representing papers, authors, and subjects. The features of papers consist of bag-of-words vectors derived from abstracts. The target node for evaluation is the paper. The edge types include paper-cite-paper, paper-ref-paper, paper-author, author-paper, paper-subject, subject-paper, paper-term, and term-paper. The node classification target is to classify papers into 3 classes: database, wireless communication, and data mining. 
    
    \item \textbf{IMDB} is a website about movies. There are 4 node types: movies, directors, actors, and keywords. A movie can have multiple class labels. The $6$ edge types include movie-director, director-movie, movie-actor, actor-movie, movie-keyword, and keyword-movie. There are 5 classes: action, comedy, drama, romance, and thriller. Movies are the targets to classify.   
    
    \item \textbf{Freebase}~\cite{43_bollacker2008freebase} is a large knowledge graph with 8 node types, including book, film, music, sports, people, location, organization, and business, and 36 edge types. The target is to classify books into 7 classes: scholarly work, book subject, published work, short story, magazine, newspaper, journal, and poem. 

    \item \textbf{PubMed} is a biomedical literature library. We use the data constructed by~\cite{45_yang2020heterogeneous}. The node types are gene, disease, chemical, and species. The $10$ edge types contain gene-and-gene, gene-causing-disease, disease-and-disease, chemical-in-gene, chemical-in-disease, chemical-and-chemical, chemical-in-species, species-with-gene, species-with-disease, and species-and-species. The target of the node classification task is to predict the disease into eight categories with class labels from ~\cite{45_yang2020heterogeneous}. The target of link prediction is to predict the existence of edges between genes and diseases. 
\end{itemize}


\subsection{Data Split}
\label{A.2 Data Split}
Following ~\cite{2_lv2021we}, we have the training and validation ratio 8:2 while keeping a fixed testing test, and the statistics of testing data, as well as training and validation data, are listed in Table \ref{tab: node_split}.

\begin{table}[ht!]
\centering
\small
\setlength{\tabcolsep}{0.9mm}
\begin{tabular}{ccccc}
\toprule
  & \#Nodes & \#Training  &\#Validation &\#Testing\\ 
 \midrule
DBLP          & 26,128  & 974  & 243  &   2,840            \\
IMDB          & 21,420  &  1,097  & 274  &  3,159  \\
ACM           & 10,942  &  726 & 181  &  2,118   \\
Freebase      & 180,098   & 1,909  & 477  & 4,446   \\
PubMed\_NC      & 63,109   &  295 & 73  &  86  \\ 
\bottomrule
\end{tabular}
\caption{Data Split of Node Classification Datasets.}
\label{tab: node_split}
\end{table}

\section{Implementation Details}


\subsection{Experimental setup}
The experiments in this study were conducted on a Linux server running Ubuntu 20.04. The server was equipped with a 13th Gen Intel(R) Core(TM) i9-13900K CPU, 128GB of RAM, and an NVIDIA GeForce RTX 4090 GPU (24GB memory). For software, we used Anaconda3 to manage the Python environment and PyCharm as the development IDE. The specific software versions were Python 3.10.14, CUDA 11.7, and PyTorch 1.13.1~\cite{44_paszke2019pytorch}. This setup provided a robust and efficient environment for our node classification experiments.

\subsection{Searched Hyperparameters}
\label{app: search hyper}

To facilitate the reproducibility of this work, we list in Table \ref{tab:my-table_parameter} the NoiseHGNN hyperparameters searched on different datasets. In addition, following the recommendations of ~\cite{2_lv2021we}, three techniques were used in the implementation of NoiseHGNN: residual connectivity, attentional residuals, and the use of hidden embeddings in the intermediate layer for the node classification task.
The search space is provided as follows:

\begin{table*}[ht!]
\centering

\resizebox{\textwidth}{!}{%
\begin{tabular}{cccccccccccc}
\toprule
\textbf{Dataset}    & $p^{A}$ & $p^{A^{\theta}}$ & k  & \multicolumn{1}{l}{Synthesizer}  & Layers & Heads & Lr   & Epochs & Patience & FeatureSize & HiddenSize \\ \midrule
\textbf{DBLP}       & 0.3     & 0.5              & 25 & GCN                                    & 2      & 8     & 5e-4 & 300    & 30       & 64          & 64         \\
\textbf{IMDB}       & 0       & 0                & 15 & MLP                                    & 5      & 8     & 5e-4 & 300    & 50       & 64          & 64         \\
\textbf{ACM}        & 0.3     & 0.4              & 35 & GCN                                    & 2      & 8     & 5e-4 & 300    & 50       & 64          & 64         \\
\textbf{PubMed\_NC} & 0.4     & 0.4              & 25 & GCN                                     & 2      & 8     & 5e-3 & 300    & 30       & 64          & 64         \\
\textbf{Freebase}   & 0.3     & 0.4              & 15 & GCN                                     & 2      & 8     & 5e-4 & 500    & 30       & 64          & 64         \\ \bottomrule
\end{tabular}%
}
\caption{Hyperparameter settings on each dataset.}
\label{tab:my-table_parameter}
\end{table*}

\begin{itemize}
    \item Learning rate lr: \{0.05, 0.005, 0.005 ,0.0005\}
    \item Weight decay : \{1e-3, 1e-4, 1e-5, 1e-6\}
    \item Number of epochs: \{300, 400, 500\}
    \item Graph generator $g_w$: \{GCN, MLP\}
    \item Number of neighbor in kNN $k$: \{5, 15, 25, 35, 45\}
    \item Edge mask rate $p^{A}$: \{0.0, 0.1, 0.2, 0.3, 0.4, 0.5, 0.6, 0.7, 0.8, 0.9\}
    \item Edge mask rate $p^{A^{\theta}}$: \{0.0, 0.1, 0.2, 0.3, 0.4, 0.5, 0.6, 0.7, 0.8, 0.9\}
    \item Early stop patience: \{30, 50\}
\end{itemize}

\subsection{Evaluation Metrics} \label{app: Evaluation Metrics}
Here we illustrate how we compute the four metrics: Macro-F1 and Micro-F1. 

\textbf{Macro-F1}: The macro F1 score is computed using the average of all the per-class F1 scores.
Denote $\text{Precision}_{c}$ as the precision of class $c$, and $\text{Recall}_{c}$ as the recall of class $c$. We have

\begin{equation}
    \begin{aligned}
        \text{Macro-F1}=\frac{1}{C}  \sum_{c\in [C]} \frac{2\text{Precision}_{c}\cdot\text{Recall}_{c}}{\text{Precision}_{c}+\text{Recall}_{c}}.
    \end{aligned}
\end{equation}

\noindent \textbf{Micro-F1}: The Micro-F1 score directly uses the total precision and recall scores. With the Precision and Recall scores of all nodes regardless of their classes, we have
\begin{equation}
    \begin{aligned}
        \text{Micro-F1}=\frac{2\text{Precision}\cdot\text{Recall}}{\text{Precision}+\text{Recall}}.
    \end{aligned}
\end{equation}

\subsection{Meta-paths Used in Baselines}
Meta-paths~ \cite{31_schlichtkrull2018modeling,32_sun2012mining} have been widely used for mining and learning with heterogeneous graphs. A meta-path is a path with a pre-defined (node or edge) types pattern, i.e., $\&P \triangleq n_1 \xrightarrow{r_1} n_2 \xrightarrow{r_2} \cdots \xrightarrow{r_l} n_{l+1}$, where $r_i\in T_e$ and $n_i\in T_v$. For instance, the ``author$\leftrightarrow$paper$\leftrightarrow$author'' meta-path defines the ``co-author'' relationship, which Can be abbreviated as APA.
Given a meta-path $\mathcal{P}$ we can re-connect the nodes in $\text{G}$ to get a meta-path neighbor graph $G_{\varphi}$. Edge $u\to v$ exists in $G_{\varphi}$ if and only if there is at least one path between u and v following the meta-path ${\varphi}$ in the original graph $\text{G}$. We provide the meta-paths used in baselines in Table~\ref{tab:meta-paths}.

\begin{table}[h!]
\centering
\small
\begin{tabular}{ccc}
\hline
Dataset & Meta-paths & Meaning \\ \hline
DBLP & \makecell[c]{APA, APTPA, \\ APVPA} & \makecell[c]{A: Author, P: Paper,\\ T: Term, V: Venue} \\ \hline

ACM & \makecell[c]{PAP, PSP,\\ PcPAP,  PcPSP, \\PrPAP,  PrPSP,\\ PTP} & \makecell[c]{P: Paper, A: Author, \\ S: Subject, T: Term, \\ c: citation relation, \\ r: reference relation } \\ \hline
Freebase & \makecell[c]{BB, BFB, \\ BLMB, BPB, \\ BPSB, BOFB, \\ BUB} & \makecell[c]{B: Book, F: Film, \\ L: Location, M: Music, \\ P: Person, S: Sport, \\ O: Organization,\\ U: bUsiness} \\ 
\hline
PubMed & \makecell[c]{DD, DGGD, \\ DCCD, DSSD} & \makecell[c]{D: Disease, G: Gene, \\ C: Chemical, S: Species} \\ \hline
\end{tabular}
\caption{Meta-paths used in our experiments.}
\label{tab:meta-paths}
\end{table}

\subsection{Algorithm} 
\label{app: Algorithm}
The training algorithm of \textbf{NoiseHGNN} is summarized in Algorithm \ref{alg:alg1}.
\begin{algorithm}[h!]
\renewcommand{\algorithmicrequire}{\textbf{Input:}}
\renewcommand{\algorithmicensure}{\textbf{Parameters:}}
\caption{The training algorithm of NoiseHGNN}
\label{alg:alg1}
\begin{algorithmic}
\REQUIRE Heterogeneous graph $G=\{\mathcal{V};\mathcal{E},\phi,\psi\}$; Feature matrix  $\mathbf{X}$; Adjacency matrix $\mathbf{A}$.
\ENSURE Number of epoch $T$;
\end{algorithmic} 
\begin{algorithmic}[1]
\STATE Generate initial parameters for all learnable parameters. 
\FOR{\text{\rm each epoch} $i=0,1,2,...,T$}
\STATE  Calculate ${Z}$ $\leftarrow$ Put ${X}$ into Eq. (\ref{eq:projection})  \\
\STATE  Calculate $\mathrm{S}$ with graph synthesizer by Eq. (\ref{eq:projection2}) and (\ref{eq:similarity})\\
\STATE  Calculate $\mathrm{A}^{\theta}$ with top-k $(\cdot)$  by Eq. (\ref{eq:adjacent}) 
\STATE Obtain augmented graph $\bar{A}$ and $\bar{A}^{\theta}$ by Eq.(\ref{eq: graph augment})

\STATE Calculate node representations $\bar{H}$,$\bar{H}^{\theta}$ with HGNN Encoder $f_{\theta}$ by Eq.(\ref{eq:correlation})-Eq.(\ref{eq: all node representations})
\STATE Calculate node projections $\bar{{\mathrm{Y}}}$,$\bar{Y}^{\theta}$ with MLP projector by Eq.(\ref{eq:objective})

\STATE Calculate $\mathcal{L}_o$ and $\mathcal{L}_s$ with cross-entropy loss by Eq.(\ref{eq: CE_LOSS})

\STATE Obtain subgraph $A^{\varphi}$ and $\hat{A}^{\theta}$ with  Eq.(\ref{eq: subgraph})

\STATE Calculate $\mathcal{L}_g$  with scaled cosine loss by Eq.(\ref{eq: Lg_LOSS})

\STATE Calculate the objective loss $\mathcal{L}$ by Eq.(\ref{eq: final objective function})
\STATE Update model parameters via gradient descent
\ENDFOR
\end{algorithmic}
\end{algorithm}

\begin{table*}[ht!]
\centering
\resizebox{\textwidth}{!}{%
\begin{tabular}{ccccccccccc}
\toprule
\textbf{} & \multicolumn{2}{c}{\textbf{DBLP}}         & \multicolumn{2}{c}{\textbf{IMDB}}         & \multicolumn{2}{c}{\textbf{ACM}}          & \multicolumn{2}{c}{\textbf{PubMed\_NC}}   & \multicolumn{2}{c}{\textbf{Freebase}}     \\ 
\midrule
          & Macro-Fl            & Micro-Fl            & Macro-F1            & Micro-F1            & Macro-Fl            & Micro-Fl            & Macro-Fl            & Micro-F1            & Macro-F1            & Micro-F1            \\ \midrule

GCN(2016)       & 90.84±0.32          & 91.47±0.34          & 57.88±1.18          & 64.82±0.64          & 92.17±0.24          & 92.12±0.23          & 9.84±1.69           & 21.16±2.00          & 27.84±3.13          & 60.23±0.92          \\
GAT(2017)       & 93.83±0.27          & 93.39±0.30          & 58.94±1.35          & 64.86±0.43          & 92.26±0.94          & 92.19±0.93          & 24.89±8.47          & 34.65±5.71          & 40.74±2.58          & 65.26±0.80          \\ 

RGCN(2018)      & 91.52±0.50          & 92.07±0.50          & 58.85±0.26          & 62.05±0.15          & 91.55±0.74          & 91.41±0.75          & 18.02±1.98          & 20.46±2.39          & -          & -          \\

RSHN(2019)      & 93.34±0.58          & 93.81±0.55          & 59.85±3.21          & 64.22±1.03          & 90.50±1.51          & 90.32±1.54          & -                   & -                   & -                   & -                   \\

HAN(2019)       & 91.67±0.49          & 92.05±0.62          & 57.74±0.96          & 64.63±0.58          & 90.89±0.43          & 90.79±0.43          & -          & -          & -          & -         \\

HetGNN(2019)    & 91.76±0.43          & 92.33±0.41          & 48.25±0.67          & 51.16±0.65          & 85.91±0.25          & 86.05±0.25          & -          & -          & -                   & -                   \\
MAGNN(2020)     & 93.28±0.51          & 93.76±0.45          & 56.49±3.20          & 64.67±1.67          & 90.88±0.64          & 90.77±0.65          & -                   & -                   & -                   & -                   \\
HGT(2020)       & 93.01±0.23          & 93.49±0.25          & 63.00±1.19          & 67.20±0.57          & 91.12±0.76          & 91.00±0.76          & 47.50±6.34          & { 51.86±4.85}          & 29.28±2.52          & 60.51±1.16          \\

simpleHGN(2021) & 94.04±0.27          & 94.46±0.20          & 63.53±1.36         & 67.36±0.57         & { 93.42±0.44}    & {93.35±0.45}    & 42.93±4.01          & 49.26±3.32          & {47.72±1.48}    & {66.29±0.45}    \\

space4HGNN(2022) &
  94.24±0.42&94.63±0.40 &
  61.57±1.19&63.96±0.43 &
  92.50±0.14&92.38±0.10 &
  45.53±4.64&49.76±3.92 &
  41.37±4.49&65.66±4.94 \\
Seq-HGNN(2023)       & \textbf{96.27±0.24}         & \textbf{95.96±0.31}       & \textbf{66.77±0.24}          & \textbf{69.31±0.27}          & \textbf{94.41±0.26}          & \textbf{93.11±0.57}          &-           & -           & -          & -        \\

SlotGAT(2024)   & \underline{94.95±0.20} & \underline{95.31±0.19} & {64.05±0.60}    & \underline{68.54±0.33} & \underline{93.99±0.23} & \underline{94.06±0.22} & \underline{47.79±3.56}    & \underline{53.25±3.40}    & \textbf{49.68±1.97} & \textbf{66.83±0.30} \\ 

\midrule
NoiseHGNN      & { 93.91±0.16}    & { 94.34±0.17}    & \underline{65.56±0.48} & { 67.85±0.30}    & 91.95±0.35          & 92.01±0.36          & \textbf{51.03±4.31} & \textbf{56.48±3.44}  & \underline{48.13±1.37 }         & \underline{66.42±0.16 }         \\ 

\bottomrule
\end{tabular}
 }
\caption{Node classification results with mean and standard deviation of Macro-F1/Micro-F1 under original dataset. Vacant
positions (“-”) mean out of memory or the original paper did not provide results. Best is in bold, and the runner-up is underlined.}
\label{tab:original_data_result}
\end{table*}

\begin{figure*}[!ht]
    \centering
    \begin{minipage}{0.33\linewidth}
        \centering
        \includegraphics[width=\linewidth]{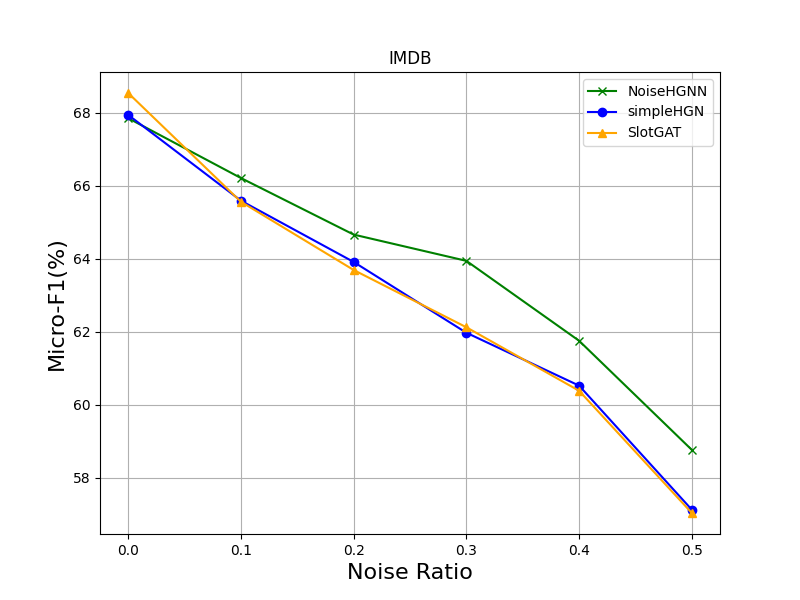}
    \end{minipage}%
    \begin{minipage}{0.33\linewidth}
        \centering
        \includegraphics[width=\linewidth]{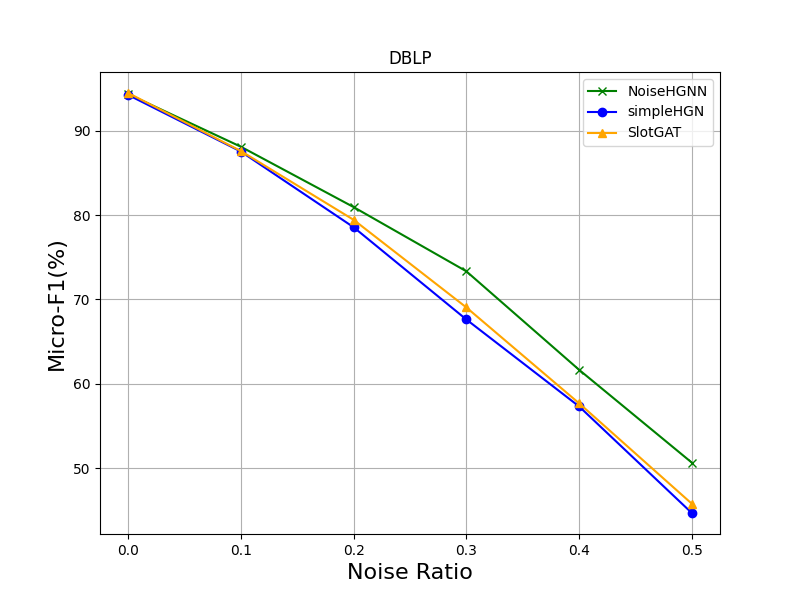}
    \end{minipage}%
    \begin{minipage}{0.33\linewidth}
        \centering
        \includegraphics[width=\linewidth]{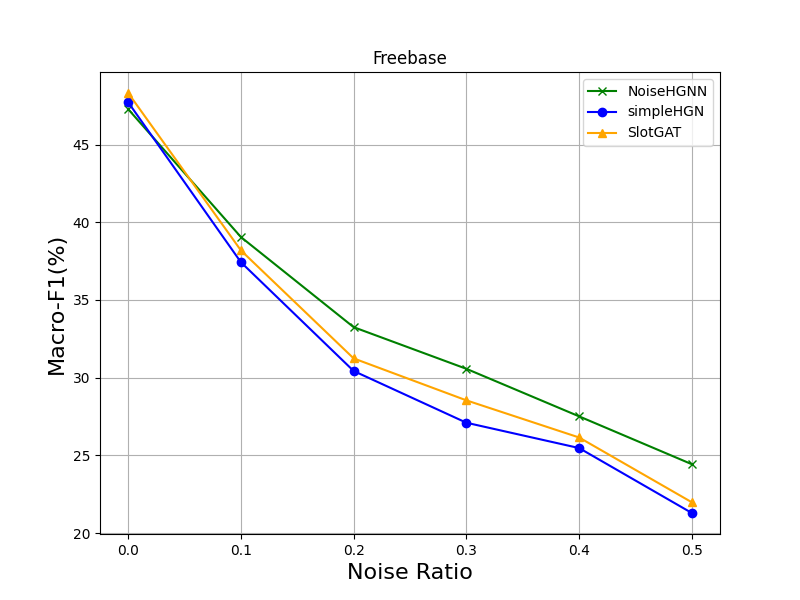}
    \end{minipage}
    \caption{Micro-F1 or Macro-F1 in the scenarios where graphs are perturbed by error link noise ratio}
    \label{fig: noise ratio}
\end{figure*}
 
\section{Additional experiments} \label{ap:Additional experiments}
 \subsection{Original Dataset Result.}  
 We also experimented with our NoiseHGNN on the original dataset (without error link perturbation). For all methods we report results from their previous work. As shown in the table \ref{tab:original_data_result}, our NoiseHGNN achieves results comparable to the state-of-the-art approach SlotGAT and Seq-HGNN. This is because our NoiseHGNN was originally designed for heterogeneous graphs with erroneous link perturbations, and thus failed to achieve the state-of-the-art results on the original dataset. Specifically, we achieved the second result on the Freebase dataset, especially on the dataset of PubMed\_NC, where our NoiseHGNN achieves 56.48\% and 51.03\% on the Micro-F1 and Macro-F1, respectively, which is far ahead of other methods. However, for the ACM dataset containing rich semantic embedding, our method cannot achieve better results, as in our previous analyses. For the results of simpleHGN, sapce4HGNN, Seq-HGNN, and SlotGAT, we reproduce their codes and use the optimal parameters to obtain their results.

\subsection{Link Prediction}
Table \ref{tab: table_LP} reports the link prediction results on LastFM and PubMed\_LP, with mean ROC-AUC and MRR and their standard deviations. Compared with SOTA's method SlotGAT, NoiseHGNN achieved comparable performance on both metrics for both datasets. This is because our NoiseHGNN was originally designed for homogeneous graphs with erroneous link perturbations. Thus failing to achieve state-of-the-art results in link prediction. Specifically, on PubMed\_LP, NoiseHGNN has an MRR of 92.35\%, 0.13\% higher than its best competitor SlotGAT, and achieved a second performance on the dataset LastFM. In terms of link prediction NoiseHGNN's performance in link prediction again shows its effectiveness.

\begin{table}[!t]
\centering
\small
\resizebox{0.99\columnwidth}{!}{
\begin{tabular}{ccccc}
\hline
 &
  \multicolumn{2}{c}{LastFM} &
  \multicolumn{2}{c}{PubMed\_LP} \\ \hline
  &
  ROC-AUC&MRR &
  ROC-AUC&MRR  \\ \hline
RGCN &
  57.21±0.09&77.68±0.17 &
  78.29±0.18&90.26±0.24 \\ 
  {DisenHAN}	&57.37±0.2&76.75±0.28&73.75±1.13&85.61±2.31 \\
HetGNN &
  62.09±0.01&83.56±0.14 &
  73.63±0.01&84.00±0.04 \\ 
MAGNN &
  56.81±0.05&72.93±0.59 &
  -  &   - \\
HGT &
  54.99±0.28&74.96±1.46 &
  80.12±0.93&90.85±0.33 \\ \hline
GCN &
  59.17±0.31&79.38±0.65 &
  80.48±0.81&90.99±0.56 \\ 
GAT &
  58.56±0.66&77.04±2.11 &
  78.05±1.77&90.02±0.53 \\ \hline
simpleHGN &
  {67.59±0.23}&{90.81±0.32} &
  {83.39±0.39}&{92.07±0.26} \\ 
space4HGNN &
  66.89±0.69&90.77±0.32 &
   81.53±2.51&90.86±1.02\\  \hline
SlotGAT &
  \textbf{70.33±0.13}&\textbf{91.72±0.50} &
  \textbf{85.39±0.28}&{92.22±0.28} \\ 

NoiseHGNN &
\underline{68.21±0.25}&\underline{91.27±0.50} &
\underline{83.79±0.28}&\textbf{92.35±0.18} \\ 
  \hline
\end{tabular}
-}
\caption{Link prediction results with mean and standard deviation of ROC-AUC/MRR.  Vacant positions (“-”) means out of memory. }
\label{tab: table_LP}

\end{table}

\subsection{Analysis of Noise Ratio}  
\label{ap: micro-f1 vs noise} 
To evaluate the robustness of NoiseHGNN under different error link noise perturbations. We modify the error link ratio from 0 to 0.5 on IMDB, DBLP, and Freebase datasets to simulate the complex real world. We compare our method with simpleHGN\cite{2_lv2021we} and SlotGAT\cite{1_zhou2023slotgat}. As shown in fig \ref{fig: noise ratio}, NoiseHGNN consistently achieves better or comparable results on both datasets. Our method exhibits a more significant performance improvement when the error link ratio becomes large, indicating that NoiseHGNN is more robust to severe error link perturbations.

\end{appendices}